\documentclass[10pt,twocolumn,letterpaper]{article}

\usepackage[pagenumbers]{cvpr} % To force page numbers, e.g. for an arXiv version

\usepackage[utf8]{inputenc} % allow utf-8 input
\usepackage[T1]{fontenc}    % use 8-bit T1 fonts
\usepackage{url}            % simple URL typesetting
\usepackage{booktabs}       % professional-quality tables
\usepackage{amsfonts}       % blackboard math symbols
\usepackage{nicefrac}       % compact symbols for 1/2, etc.
\usepackage{microtype}      % microtypography
\usepackage{amsthm}
\usepackage{multicol}
\usepackage{multirow}
\usepackage{algorithm}
\usepackage{algorithmic}
\usepackage{longtable}
\usepackage{booktabs}
\usepackage{extarrows}
\usepackage{textcomp}
\usepackage{booktabs}
\usepackage{graphicx}
\usepackage{enumitem}
\usepackage{soul}
\usepackage{threeparttable}

\usepackage{colortbl}
\newtheorem{theorem}{Theorem}

\newtheorem{proposition}[theorem]{Proposition}
\newtheorem{definition}{Definition}

\usepackage[dvipsnames]{xcolor}

\definecolor{cvprblue}{rgb}{0.21,0.49,0.74}
\usepackage[pagebackref,breaklinks,colorlinks,citecolor=cvprblue]{hyperref}

\def\paperID{2810} % *** Enter the Paper ID here
\def\confName{CVPR}
\def\confYear{2024}

\title{Rethinking Classifier Re-Training in Long-Tailed Recognition: \\
A Simple Logits Retargeting Approach}

\author{Han Lu$^{1,*}$,
Siyu Sun$^{1,*}$,
Yichen Xie$^{2}$, Liqing Zhang$^{1}$, Xiaokang Yang$^{1}$, Junchi Yan$^{1,\dag}$ \\
$^{1}$ Department of Computer Science and Engineering, Shanghai Jiao Tong University \\ $^{2}$ University of California, Berkeley
\\
$^*$ Equal contributions \quad $^\dag$ Correspondence author
}

\begin{document}
\maketitle
\begin{abstract}
In the long-tailed recognition field, the Decoupled Training paradigm has demonstrated remarkable capabilities among various methods. This paradigm decouples the training process into separate representation learning and classifier re-training. Previous works have attempted to improve both stages simultaneously, making it difficult to isolate the effect of classifier re-training. Furthermore, recent empirical studies have demonstrated that simple regularization can yield strong feature representations, emphasizing the need to reassess existing classifier re-training methods. In this study, we revisit classifier re-training methods based on a unified feature representation and re-evaluate their performances. We propose a new metric called Logits Magnitude as a superior measure of model performance, replacing the commonly used Weight Norm. However, since it is hard to directly optimize the new metric during training, we introduce a suitable approximate invariant called Regularized Standard Deviation. Based on the two newly proposed metrics, we prove that reducing the absolute value of Logits Magnitude when it is nearly balanced can effectively decrease errors and disturbances during training, leading to better model performance. Motivated by these findings, we develop a simple logits retargeting approach (LORT) without the requirement of prior knowledge of the number of samples per class. LORT divides the original one-hot label into small true label probabilities and large negative label probabilities distributed across each class. Our method achieves state-of-the-art performance on various imbalanced datasets, including CIFAR100-LT, ImageNet-LT, and iNaturalist2018.

\end{abstract}    
\section{Introduction}

Real-world datasets often exhibit long-tailed distributions ~\cite{buda2018systematic, zhang2021deep}. The majority of the samples belong to a few major classes, while the remaining samples are distributed across many rare classes. Although traditional classification methods~\cite{he2016deep, redmon2017yolo9000} perform well on balanced datasets, they may not be effective in such imbalanced datasets since they tend to prioritize the majority classes and pay less attention on learning the minority classes~\cite{buda2018systematic, he2009learning, van2017devil}. Long-tail recognition (LTR) has emerged as a challenging and important task, where models are trained on imbalanced datasets and then evaluated on well-balanced datasets~\cite{zhang2021deep}.

Various techniques have been developed for LTR, including class re-weighting~\cite{cui2019class, lin2017focal, ren2020balanced}, class re-sampling~\cite{kang2019decoupling, mahajan2018exploring, ren2020balanced, wang2020devil}, and others. Among these techniques, the Decoupled Training paradigm~\cite{kang2019decoupling} has shown remarkable efficacy. This approach divides the training process into two different stages: representation learning and classifier retraining.  In the first stage, the model is trained to learn a general feature representation without considering class imbalance. In the second stage, the classifier is retrained using the acquired feature representation, with an emphasis on minority classes.

However, previous works~\cite{alshammari2022long, desai2021learning, ren2020balanced, zhong2021improving} on Decoupled Training have often focused on simultaneously improving both the representation learning and classifier retraining stages, which may obscure the impact of classifier retraining based on different feature representations. LTWB~\cite{alshammari2022long} has demonstrated that simple regularization techniques can produce strong feature representations in the first stage, outperforming the original first-stage representation learning methods. These findings highlight the need for a rigorous evaluation and assessment of classifier retraining methods. By carefully evaluating the effectiveness of different classifier retraining methods based on unified feature representations, one can better identify the truly effective factors and make possible improvements.

Therefore, we revisit classifier re-training methods, including re-weighting~\cite{kang2019decoupling, cao2019learning, ren2020balanced, lin2017focal, cui2019class}, re-sampling~\cite{shen2016relay}, parameter regularization~\cite{kang2019decoupling, alshammari2022long}, and direct logits modification during inference~\cite{kang2019decoupling, menon2020long}. To better understand the similarities and differences between these methods, we design a unified formulation to provide a comprehensive overview. Using the same strong first-stage feature representations, we conduct experiments to compare the accuracy of these methods. Notably, the performance of re-evaluated methods shown in Tab.~\ref{tab:long-tailed-loss} far exceeds those claimed in previous literature, providing a new fair benchmark for further analysis. 

% \red{Among these methods, class-balanced loss~\cite{ren2020balanced} applied with  binary cross entropy loss obtains the best performance. We contribute the success to considering negative-class predictions in addition to ground truth labels, as opposed to solely focusing on the latter.}

Based on the above experimental results, we conduct further analyses. Previous works~\cite{alshammari2022long,kang2019decoupling} use weight norm of the classifier as a measure to demonstrate that their models do not overfit to the majority classes. However, this metric is not always reliable as it disregards the feature magnitude for different classes and limits the model’s representational capacity. Instead, we propose a novel evaluation metric called Logits Magnitude. This metric considers the distribution of logits directly and is defined as the difference between the mean logits of positive and negative samples for each class. The performance of different models exhibits a positive correlation with the balance of Logits Magnitude between classes. However, directly optimizing this new metric during training is not feasible. To address this challenge, we propose a solution by introducing an approximate invariant called Regularized Standard Deviation. This invariant is calculated as the standard deviation divided by Logits Magnitude. By analyzing the relationship between this invariant and Logits Magnitude, we demonstrate that reducing the absolute value of Logits Magnitude, particularly when it is nearly balanced, can effectively decrease errors and disturbances during the training process. This finding inspires our method.

\begin{figure}[t!]
    \centering
    \begin{subfigure}{\linewidth}
        \caption{\textbf{Vanilla Cross-Entropy (CE).} "Many" class logit $2.58$ is larger than the ground truth "Few" class logit $1.69$, leading to the misclassification.}
        \label{fig:Logits_CE}
        \includegraphics[width=\linewidth]{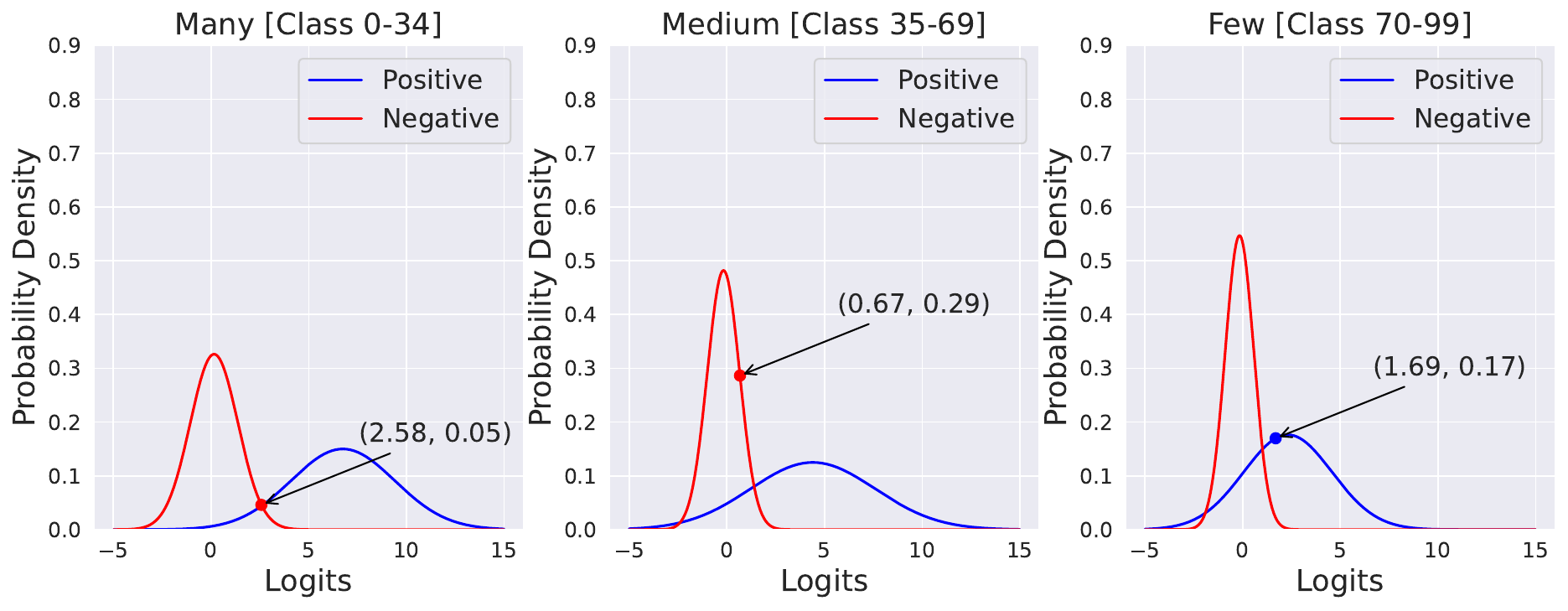}
    \end{subfigure}
    \begin{subfigure}{\linewidth}
        \caption{\textbf{LORT (ours).} "Few" class logit $1.57$ outperforms the other  classes.}
        \label{fig:Logits_Ours}
        \includegraphics[width=\linewidth]{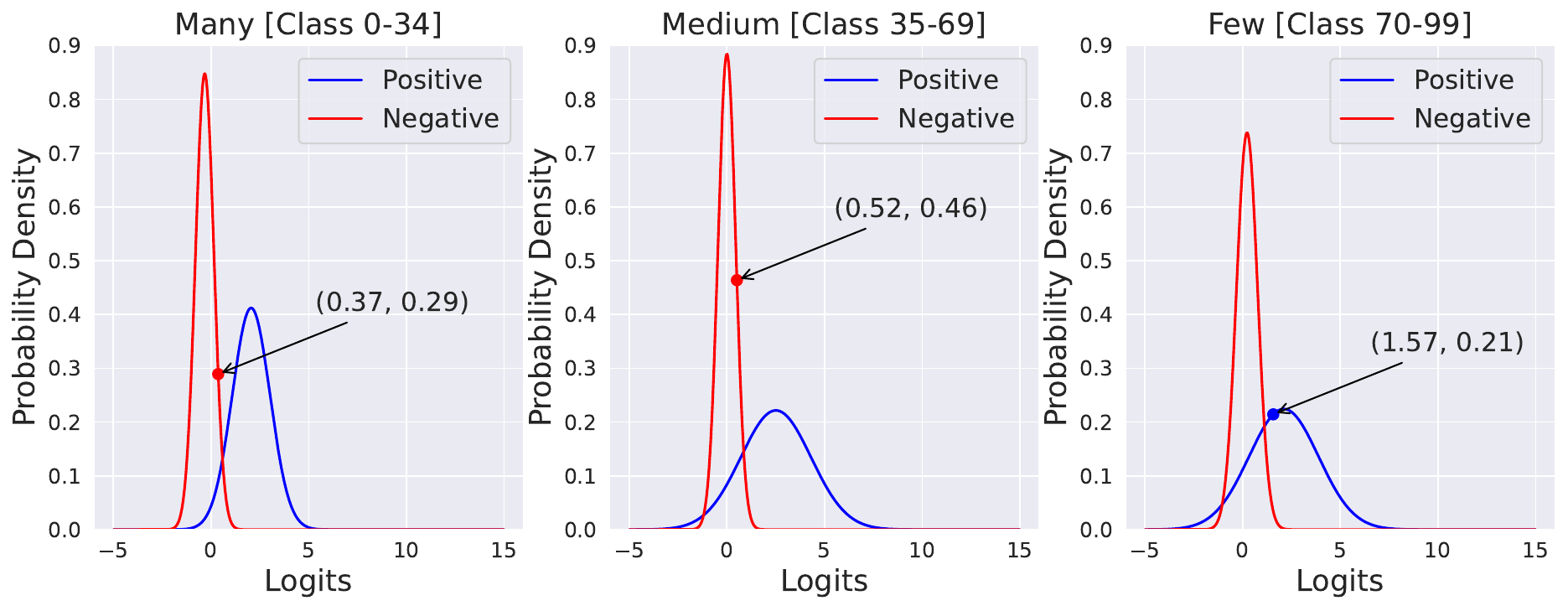}
    \end{subfigure}
    % \vspace{-10pt}
    \caption{\textbf{An Intuitive View of Logits Magnitude Influence.} The two models are trained on CIFAR100-LT with IR=100. We take the instance indexed $7202$ in test set as an example. Its true class is $75$, but it is erroneously classified as class $3$ in Many class by Vanilla CE . The blue points represent the logits of the true positive class, while the red points correspond to the logits of the negative class. Our proposed method, LORT, can obtain the more balanced and smaller logits magnitudes (defined as the difference between the positive logits and negative logits of the same class), helping discrimination in a multi-class classification scenario.
    % The classification performance between positive and negative samples within its single class remains almost unchanged between these two methods. However, in a multi-class classification scenario, the interference between classes is mitigated.
    }
    \vspace{-10pt}
    \label{fig:Logits_CE_Ours_Intro}
\end{figure}

Hence, we propose a simple logits retargeting approach called Logits Retargeting (LORT). LORT does not require prior knowledge of the number of samples per class and can effectively reduce the logits magnitude while maintaining balance across classes. It divides the original one-hot label into small true label probabilities and large negative label probabilities distributed to each class. We demonstrate, both theoretically and experimentally, that a larger negative label probability can lead to better results. Fig.~\ref{fig:Logits_CE_Ours_Intro} provides an intuitive explanation for the advantages of our LORT approach.  When considering each class individually, the discriminative power between positive and negative samples of the same class remains relatively constant, which we attribute to an inherent error in dataset and representation learning. However, in a multi-class classification scenario, LORT lessens the influence of predictions from other classes. This elimination of the superiority of majority classes enhances the discernibility of each sample.
Our simple method achieves state-of-the-art performance on various mainstream imbalanced datasets, such as CIFAR100-LT, ImageNet-LT, and iNaturalist2018. Extensive ablation studies also verify the stability of our method against hyperparameter variation. Meanwhile, it can serve as a plug-and-play retraining method that can effectively enhance previous methods.

\textbf{Our contributions are as follows:}
\textbf{1)} We revisit previous classifier retraining methods for long-tailed recognition and update their experimental results to enable fair comparisons and further analysis.
\textbf{2)} We propose two metrics, namely "Logits Magnitude" and "Regularized Standard Deviation," which effectively compare the distributions of logits values across different methods and classes. These metrics provide insights into the model's performance and highlight the conditions for achieving improved performance.
\textbf{3)} We introduce a simple logits retargeting approach called Logits Retargeting (LORT) that does not require prior knowledge of the number of samples per class. LORT mitigates the influence of overfitting and bias in the training set, achieving state-of-the-art performance on three commonly used datasets.

% \vspace{-15pt}
\section{Related Work}
% \vspace{-10pt}
\begin{table*}[tb!]
    \renewcommand{\arraystretch}{1.2}
  \centering
      \caption{Study of the classifier re-training methods in decoupled learning scheme in LTR. Vanilla FC trains a general representation, performing itself in a role as both the baseline and the backbone feature extractor. Experiments are conducted on CIFAR100-LT with imbalanced ratio 100 and  constant $\gamma, \beta, \delta, \tau$. See details for the listed formulation in Sec.~\ref{sec:revisit_crt}.}
       \vspace{-5pt}
  \resizebox{\linewidth}{!}{
    \begin{tabular}{c|c|cccc}
        \toprule
        \multirow{2}{*}{\textbf{Methods}} & \multirow{2}{*}{\textbf{Formulation}} & \multicolumn{4}{c}{\textbf{ Accuracy}} \\
        % \cline{3-6}
        & & \textbf{All} & \textbf{Many} & \textbf{Medium} & \textbf{Few} \\
        \hline
        Vanilla FC & $g_i = \mathbf{W}_i^\top f(\mathbf{x})$ & 48.44 & \textbf{78.74} & 47.37 & 14.33 \\
        \hline
        Vanilla $\operatorname{Cosine}$ & $g_i= \mathbf{W}_i^\top / ||\mathbf{W}_i^\top|| \cdot f(\mathbf{x}) / ||f(\mathbf{x})|| $ & 51.66 & 77.77 & 49.60 & 23.60\\
        \hline
        Learnable Weight Scaling~\cite{kang2019decoupling} & $g_i = \mathbf{c}_i \cdot \mathbf{W}_i^\top f(x)$, where $\mathbf{c}_i$ is learnable & 48.99 & 78.51 & 47.51 & 16.26 \\
        \hline
        LDAM Loss~\cite{cao2019learning} & $g_i = \mathbf{W}_i^\top f(\mathbf{x}) - \mathbf{1}\{i=y\} \cdot C / n_i^{\gamma}$ & 52.94 & 72.37 & 54.31 & 28.67 \\
        \hline
        Balanced-Softmax Loss~\cite{ren2020balanced} & $g_i = \mathbf{W}_i^\top f(\mathbf{x}) + \log \left(n_i \right)$ & 52.23 & 77.63 & 50.74 & 24.33 \\
        \hline
        Re-Sampling~\cite{shen2016relay} & $r_s(i) \propto 1 / n_i$, $r_s(i)$ is the sample ratio of class $i$ & 48.44 & 78.74 & 47.37 & 14.33 \\
        \hline
        Focal Loss~\cite{lin2017focal} & $r(y)=\left(1 - \mathbf{p}_y\right)^\gamma$, $\mathbf{p}_y$ is the predicted probability of $y$ & 50.94 & 78.00 & 49.97 & 20.50 \\     
        \hline
        Class-Balanced Loss~\cite{cui2019class} & $r(y)=(1-\beta) /\left(1-\beta_y^n\right)$ applied with BCE loss & \textbf{54.30} & 71.71 & \textbf{51.25} & \textbf{37.53} \\
        \hline 
        MaxNorm~\cite{alshammari2022long} & $||\mathbf{W}_i||^2_2 \leq \delta^2$  & 51.64 & 77.60 & 49.37 & 23.99\\  
        \hline
        $\tau$-normalized~\cite{kang2019decoupling}  & $h_i=\left(\mathbf{W}_i /\left\|\mathbf{W}_i\right\|^\tau\right)^\top f(\mathbf{x})$ & 53.01 & 74.74 & 47.31 & 34.30 \\  
        \hline
        Post-hoc Logit Adjustment~\cite{menon2020long}  & $h_i = \mathbf{W}_i^\top f(\mathbf{x})-\tau \log \left(n_i\right)$ & 52.45 & 75.57 & 47.71 & 31.00 \\

        \bottomrule
    \end{tabular}
    }
    \vspace{-10pt}
    \label{tab:long-tailed-loss}%
\end{table*}%

\textbf{Long-Tailed Recognition (LTR).} In real-world scenarios, the collected data exhibits a long-tail distribution, wherein a small fraction of classes possess a substantial number of samples and the remaining classes are associated with only a few samples~\cite{zhang2021deep}.
% Long-Tailed Recognition (LTR) encompasses the utilization of imbalanced training data, mirroring the real-world distribution, and necessitates models to accurately predict the balanced test data to ensure fairness. 
To address long-tailed class imbalance and the failure of the vanilla empirical risk minimization methods,  massive studies have been conducted recently. (i) \emph{Re-sampling} aims to balance the training samples of different class~\cite{kang2019decoupling, wang2020devil, mahajan2018exploring, ren2020balanced} by over-sampling the data from the tail class and under-sampling the data from the head class. (ii) \emph{Cost-sensitive learning} adjusts loss values for different classes to make a balance, such as focal loss ~\cite{lin2017focal}, class-balanced loss (CB) ~\cite{cui2019class}, balanced softmax loss (BS) ~\cite{ren2020balanced}, LADE~\cite{hong2021disentangling}  and 
%soft margin loss extended by label-distribution-aware margin 
LDAM ~\cite{cao2019learning}. (iii) \emph{Representation Learning} focuses on improving the feature extractors, consisting some paradigms, i.e., metric learning~\cite{huang2016learning,zhang2017range}, prototype learning~\cite{liu2019large, zhu2020inflated}, and transfer learning~\cite{zhong2021improving,li2021self}.  (iv) \emph{Classifier Design
} replaces the linear classifier  
with well-designed classifiers, such as scale-invariant cosine classifier~\cite{wu2021adversarial} and $\tau$-normalized classifier~\cite{kang2019decoupling}. 
(v) \emph{Decoupled Training} decouples the learning into representation learning and classifier retraining, rather than training them jointly. 
(vi) Other methods use data augmentation~\cite{zada2022pure, zhong2021improving, perez2017effectiveness,shorten2019survey,chou2020remix}, self-supervised learning~\cite{liu2021self, cui2021parametric}, multi-expert ensemble \cite{cai2021ace, cui2022reslt,zhang2021test} and so on~\cite{zhang2021deep}. These models require more data or computation resources and are inclined to achieve high accuracy. In the paper, we report them as the upper bound baseline.

\textbf{Decoupled Training.} The Decoupled Training scheme \cite{alshammari2022long, cui2019class,zhang2021distribution, ren2020balanced,zhong2021improving} in LTR separates the joint learning process into feature learning and classifier learning stages. 
Decoupling \cite{kang2019decoupling} is the pioneering work to experimentally evaluate different sampling strategies in the first stage and classifier design (e.g. cRT and  $\tau$-norm) in the second stage. KCL~\cite{kang2021exploring} empirically finds a balanced
feature space important for LTR. MiSLAS~\cite{zhong2021improving} empirically observes data mixup benefit feature extractor in the first stage but will do harm to classifier performance in the second stage. DisAlign ~\cite{zhang2021distribution} develops the distribution alignment strategy with a generalized re-weight method. 
% The current state-of-the-art method Long-Tailed Recognition via Weight Balancing
(LTWB)~\cite{alshammari2022long} emphasizes parameter regularization in both feature learning and classifier finetuning, achieving state-of-the-art performance. 
% Modifying the weight decay parameter to prevent overfitting can yield a substantial improvement in the performance of the initial stage. The robust and generalizable first-stage feature extractor surpasses prior feature extraction methods, enabling us to focus primarily on finetuning the second-stage classifier. Consequently, we reassess the current classifier re-training method based on the powerful feature extractor and provide a novel analysis from a fresh perspective, culminating in our proposal of a simple yet effective method.

\section{Rethinking Classifier Re-Training}

\subsection{Revisiting Classifier Re-Training methods}
\label{sec:revisit_crt}
% For long-tailed recognition, \cite{kang2019decoupling} introduces the Decoupling Training Paradigm to effectively improve the accuracy. 
% Its training procedure consists of two steps. Firstly, the network is trained by cross-entropy (CE) loss while considering the unbalanced dataset in the usual manner. Secondly, the classifier is re-trained or fine-tuned alone to adapt to the unbalanced distribution. 
For long-tailed recognition, Decoupled Training Paradigm ~\cite{kang2019decoupling} can effectively improve the performance. Various training paradigms for classification heads have been explored, including both joint training from scratch~\cite{cao2019learning, ren2020balanced, shen2016relay} and fine-tuning only~\cite{kang2019decoupling, lin2017focal, cui2019class, alshammari2022long, menon2020long}. However, it still lacks systematic studies to find the most appropriate method for classifier re-training, as previous studies~\cite{alshammari2022long, cui2019class, ren2020balanced, zhong2021improving} often involve different contributions using various backbone feature extractors. 
%  Recently, \cite{alshammari2022long} proposed a backbone training method called LTWB, which involves adjusting the hyper-parameter weight decay, and  yields robust feature representations in the first stage. 
To provide a fair comparison, we conduct a comparative analysis of these methods using the same strong feature representations trained in first stage by LTWB~\cite{alshammari2022long}. During the re-training process, we freeze the weights of the backbone and train the classification head only.

\textbf{Notation.} Suppose there are $K$ classes with $n_i$ denoting the sample number of class $i$. The deep feature extracted from an image $\mathbf{x}$ is represented by $f(\mathbf{x})$, while the classifier weight vectors are denoted as $\mathbf{W} = [\mathbf{W}_1, ..., \mathbf{W}_K]$.

Inspired by \cite{wu2021adversarial}, we construct a unified formula for the different methods:
\begin{equation}
    \begin{aligned} 
    \mathcal{L} (\mathbf{W} ; f(\mathbf{x}), y) & = -r_w(y) \cdot \log \left(\frac{e^{\mathbf{z}_y}}{\sum_i e^{\mathbf{z}_i}}\right), \\
    \text {  where } \mathbf{z}_i & = g_i\left(\mathbf{W}_i, f(\mathbf{x})\right).
    \label{equ:formulation}
    \end{aligned}
\end{equation}
Given input feature $f(\mathbf{x})$ and the label $y$, $r_w(y)$ denotes the re-weighting factor for current label categories and $\mathbf{z}_i = g_i(\mathbf{W}_i, f(\mathbf{x}))$ represents the calculated logit after classifier. The common used classifiers like Vanilla FC are $g (\mathbf{W}, f(\mathbf{x})) = \mathbf{W}^\top f(\mathbf{x}) + \mathbf{b}$. For simplicity, we omit the bias $\mathbf{b}$ in the table. The methods Vanilla Cosine, Learnable Weight Scaling (LWS)~\cite{kang2019decoupling}, Label-Distribution-Aware Margin Loss~\cite{cao2019learning}, Balanced-Softmax Loss~\cite{ren2020balanced} focus on designing different classifier function $g$. Re-Sampling~\cite{shen2016relay} guarantees the same sampling probability of each class and we use $r_s(i)$ as the sampling ratio, acting as the supplementary of the equation.
Focal Loss~\cite{lin2017focal} and Class-Balanced Loss~\cite{cui2019class} design the class weights $r_w(y)$ according to class label $y$. 
In addition to the general formulation in Eq.~\ref{equ:formulation}, there are other perspectives. MaxNorm~\cite{alshammari2022long} regularizes the weight norm through Projected Gradient Descent (PGD) in practice. $\tau$-normalized~\cite{kang2019decoupling} and  Logit Adjusted Loss~\cite{menon2020long} use $h_i$ as the logit modification function in the inference stage. We conduct the experiments on CIFAR100-LT dataset with imbalanced ratio 100 and report the results in Tab.~\ref{tab:long-tailed-loss}. Based on the statistical results and theoretical analysis, we propose a better performance measurement Logits Magnitude.
% Based on the formulation and experimental data of these methods, we propose a better performance measurement.

\subsection{Logits Magnitude} \label{sec:conjection}
 
Consider the general form of CE without regularization, assuming the label vector as $\mathbf{y}$ instead of one-hot label in Eq.~\ref{equ:formulation}.
\begin{equation}
    \mathcal{L}(\mathbf{W}, \mathbf{b}; f(\mathbf{x}), \mathbf{y}) = \sum_i -\mathbf{y}_i \cdot \log \left(\frac{e^{\mathbf{z}_i}}{\sum_j e^{\textbf{z}_j}}\right), \\ 
\end{equation}
where $\mathbf{z} = \mathbf{W}^\top f(\mathbf{x}) + \mathbf{b}$ and define $\mathbf{s}$ as the final predicted probability that $\mathbf{s}_i = e^{\mathbf{z}_i} / \sum_j e^{\textbf{z}_j}$.

\begin{proposition}[\textbf{Bias Convergence}]
Given a deterministic matrix $\mathbf{W}$, loss function $\mathcal{L}$ is a convex function with respect to $\mathbf{b}$ and will reach the global minimum as the corresponding bias $\mathbf{b}$ converges. 

\begin{proof}
Calculate the Hessian matrix of $\mathcal{L}$ w.r.t. $\mathbf{b}$ as $\mathbf{H}$
\begin{equation}
    \mathbf{H}_{ij}=
    \begin{cases}
        \mathbf{s}_i(1 - \mathbf{s}_i) &\text{if $i = j$} \\
        -\mathbf{s}_i \mathbf{s}_j &\text{if $i\neq j$}
    \end{cases}
\end{equation}

The Hessian matrix $\mathbf{H}$ is a positive semi-definite matrix because $\mathbf{x}\mathbf{H}\mathbf{x}^\top \geq 0$ for any $\mathbf{x}$. (Details in Appendix~\ref{app:proof}).
\end{proof}
\end{proposition}

% \begin{figure}[tb!]
%     \centering
%     \includegraphics[width=\linewidth]{figure/Logits_Magnitude.pdf}
%     \caption{\textbf{Overview of Logits Magnitude} for various methods on CIFAR100-LT with an imbalanced ratio of 100. Classes are grouped into segments of 10, and mean values are computed for comparative analysis. The methods in the legend are sorted in ascending order of performance from top to bottom. The difference in means between positive and negative samples is evaluated for each class in the test set. Magnitude regularization using the 1-norm is employed to enhance comparability.}
%     \label{fig:Logits_Magnitude}
%     \vspace{-10pt}
% \end{figure}
 
Therefore, our focus shifts to the optimization problem concerning the weight matrix $\mathbf{W}$. In all cross-entropy based methods, the aforementioned conclusion remains valid. 

The precise values of individual vectors $\mathbf{W}_i$ do not influence the classification accuracy. Scaling both $\mathbf{W}$ and $\mathbf{b}$ by the same factor enables us to modify $\mathbf{W}$ without impacting the final predictions. As a result, the relative magnitude among different classes becomes the only critical factor.
%for accuracy.

Previous techniques such as MaxNorm~\cite{alshammari2022long} and $\tau$-Norm~\cite{kang2019decoupling} have been employed to constrain the length of each vector in $\mathbf{W}$. However, these Weight Norm constraint methods can not effectively characterize the relationships between different classes. Considering the equation $\mathbf{z}_i=\mathbf{W}_i^\top f(\mathbf{x}) + \mathbf{b}_i$, it is important to note that the difference in the distribution of the feature $f(\mathbf{x})$ itself and its correlation with different classes can also influence this relative relationship. Therefore, we propose a more reasonable metric directly from the perspective of logits and compare it with the commonly used Weight Norm as follows:

\begin{figure}[t!]
    \centering
    \includegraphics[width=\linewidth]{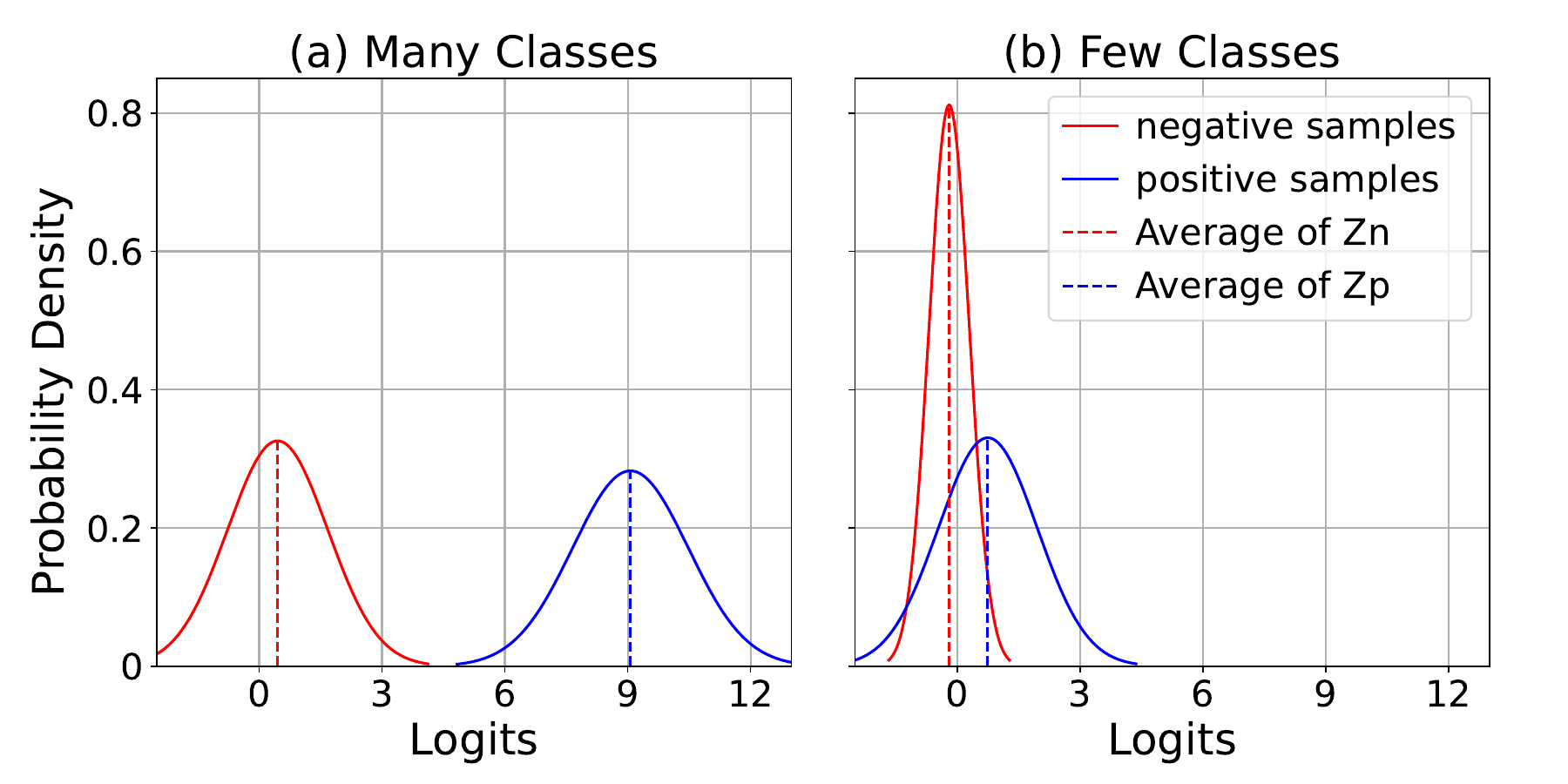}
    \vspace{-10pt}
    \caption{\textbf{Visual interpretation of Logits Magnitude.} The Logits Magnitude L is defined as the difference between the mean logits of positive (dashed blue line) and negative (dashed red line) samples.}
    \label{fig:Practical_Interpretation_main}
    \vspace{-15pt}
\end{figure}

\begin{definition}[\textbf{Logits Magnitude}]
\label{def:Logits_Magnitude}
Suppose the number of classes is $K$, $\mathbf{z}_{Pi}$ represents the logits of positive samples of class $i$ and $\mathbf{z}_{Ni}$ represents the logits of the negative samples where negative samples refer to all samples that do not belong to this class $i$. The Logits Magnitude $\mathbf{L} \in \mathbb{R}^{K}$ is defined as the difference between the mean logits of positive and negative samples for each class.
\begin{equation}
    \mathbf{L_i} = \mathbb{E}[\mathbf{z}_{\text{P}i}] - \mathbb{E}[\mathbf{z}_{\text{N}i}].
\end{equation}
\end{definition}

Fig.~\ref{fig:Practical_Interpretation_main} provides an intuitive visual interpretation of Logits Magnitude. We compute  the mean and variance of the logits for the positive and negative classes, and fit them to a normal distribution. We display schematic diagrams for the "Many" and "Few" classes as a whole for comparison.

%Fig.~\ref{fig:Practical_Interpretation} displays the schematic diagrams for the "Many" and "Few" class, where we observe that the "Many" category is easier to distinguish, while the "Few" category is more challenging. Additionally, we notice that the "Many" category exhibits lower variance, whereas the "Few" category has higher variance, which aligns with our intuition. Building upon these observations, we propose the concept of Regularized Standard Deviation to better analyse.

\begin{proposition}
[\textbf{Arbitrary Vector Length}]
\label{prop:vector_length}
If there exist optimal parameters $\mathbf{W}^*$ and $\mathbf{b}^*$ such that convergence is achieved in the train set, then there exist a series of convergence points $(\mathbf{W}', \mathbf{b}^*)$ where $\mathbf{W}'$ can have arbitrary magnitudes. 

Let $\mathbf{W}_i' = \mathbf{W}^*_i + \varepsilon$, we have 
\begin{equation}
    \begin{aligned}
        \mathbf{s}_i' = \frac{e^{\mathbf{z}'_i}}{\sum_j{e^{\mathbf{z}'_j}}} & = \frac{1}{\sum_j{e^{\mathbf{W}'_jf(\mathbf{x}) + \mathbf{b}^*_j - \mathbf{W}'_i f(\mathbf{x})-\mathbf{b}^*_i}}} \\
        & = \frac{1}{\sum_j{e^{\mathbf{z}_j - \mathbf{z}_i}}} = \mathbf{s}_i,
    \end{aligned}
\end{equation}
\end{proposition}
The predicted probability is not affected by $\varepsilon$.
Suppose $\varepsilon$ is a random variable and $\mathbb{E}[\varepsilon]=0$
\begin{equation}
    \mathbb{E}[||\mathbf{W}'_i||_2^2] = ||\mathbf{W}_i^*||_2^2 + \mathbb{E}[||\varepsilon||_2^2],
\end{equation}
but 
\begin{equation}
    \mathbb{E}[\mathbf{L}_i'] = \mathbf{L}^*_i.
\end{equation}

\begin{figure*}[t]
    \centering
    \begin{subfigure}{0.49\linewidth}
        \caption{Logits Magnitude}
        \label{fig:Logits_Magnitude}
        \includegraphics[width=\linewidth]{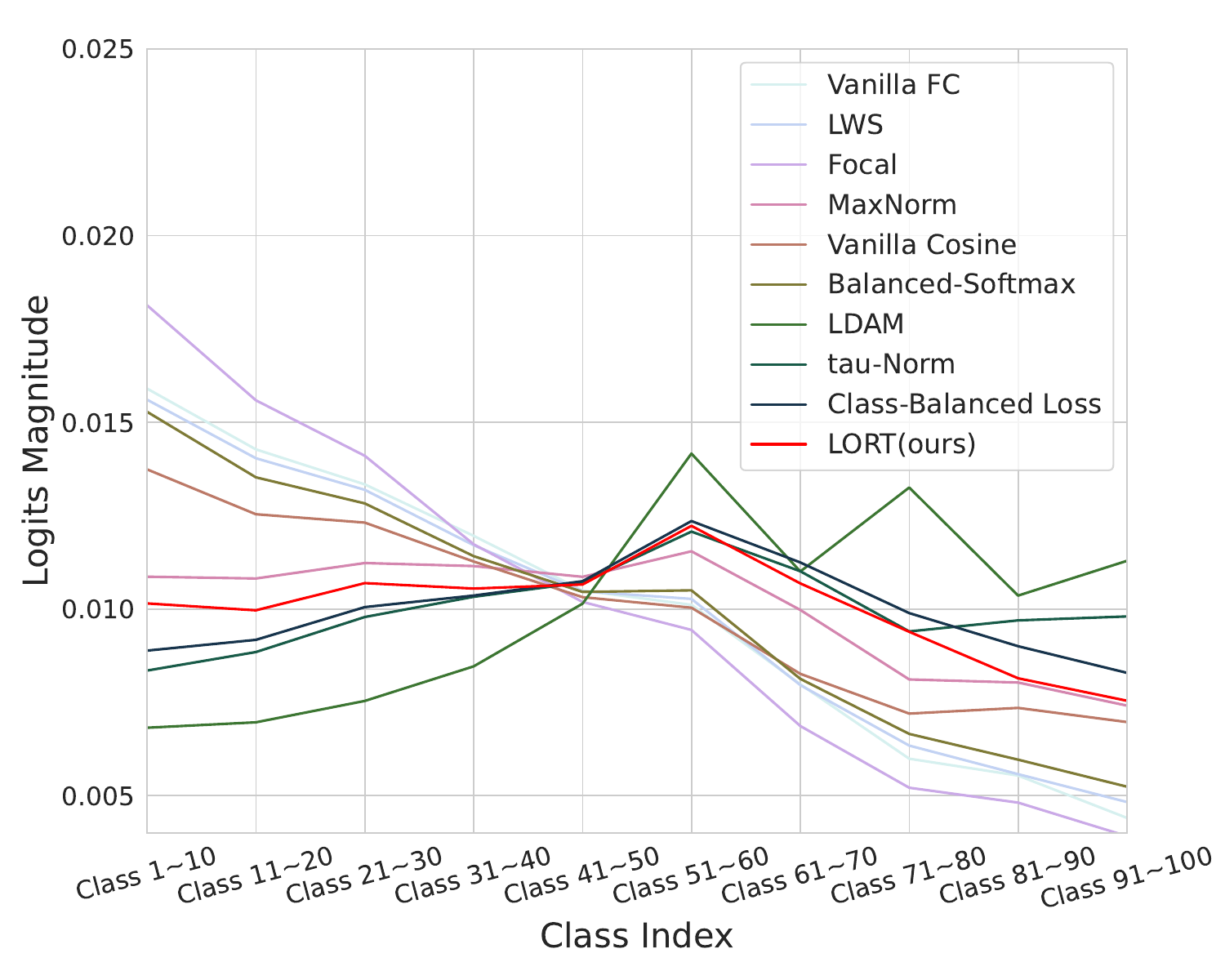}
    \end{subfigure}
    \hspace{0.01in}
    \begin{subfigure}{0.49\linewidth}
        \caption{Regularized Standard Deviation}
        \label{fig:Regularized_Standard_Deviation}
        \includegraphics[width=\linewidth]{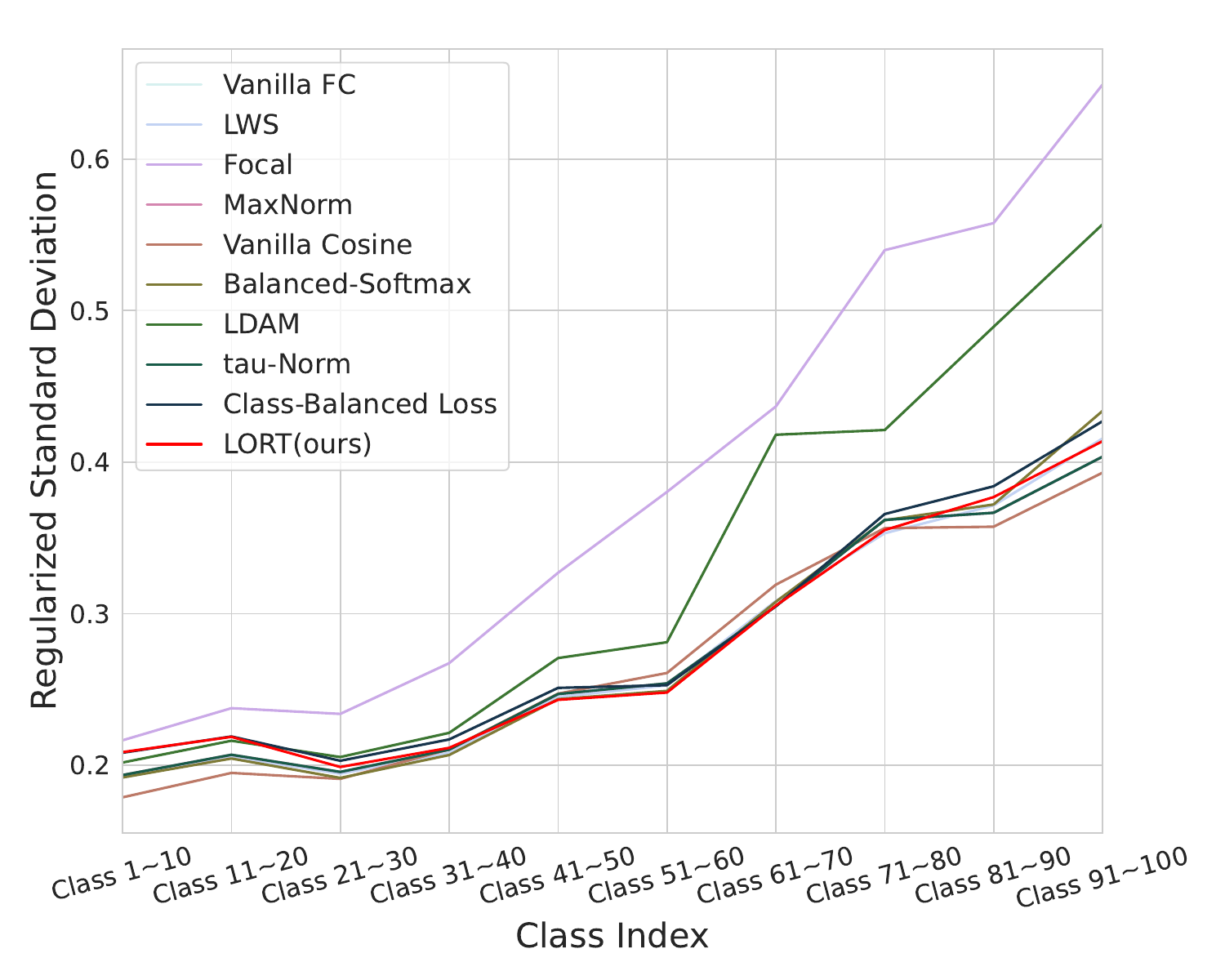}
    \end{subfigure}
    \vspace{-10pt}
    \caption{Overview of our proposed metrics. \textbf{Left: Overview of Logits Magnitude} for various methods on CIFAR100-LT with an imbalanced ratio of 100. Classes are grouped into segments of 10, and mean values are computed for comparative analysis. \textbf{The methods in the legend are sorted in ascending order of performance from top to bottom.} The difference in means between positive and negative samples is evaluated for each class in the test set. Magnitude regularization using the 1-norm is employed to enhance comparability. \textbf{Right:} \textbf{Overview of Regularized Standard Deviation}. The true distribution of logits $\mathbf{z}$ for each class is not available considering the lack of the samples, so the displayed results represent computations on the test set. % \blue{The trends align consistently with the results computed separately for positive and negative samples.}
    }
    
 \label{fig:Overview_metric}
    \vspace{-10pt}
\end{figure*}

Proposition~\ref{prop:vector_length} highlights that the magnitude of weight vectors can be arbitrary without affecting optimal prediction. This suggests that direct measures of the weight matrix, like weight norm, are unreliable and limit the model’s capability. 

Previous methods such as \cite{alshammari2022long, kang2019decoupling} often incorporate regularization as a crucial parameter. Regularization helps mitigate the influence of $\varepsilon$, but also decreases the capabilities of $\mathbf{W}$. In comparison, Logits Magnitude exhibits superior invariance.

Logits Magnitude is effective in characterizing the performance. As observed in Fig.~\ref{fig:Logits_Magnitude} where the methods are listed in increasing order of performance, \textbf{we have empirically found that achieving a more uniform distribution of Logits Magnitude tends to yield improved performance}. 
Fig.~\ref{fig:Logits_CE_Ours_Intro} illustrates the intuitive advantage of balanced Logits Magnitude on performance, particularly for the few classes.
% An intuitive view is provided in Appendix~\ref{app:logitsMagnitude} in terms of the practical interpretation and balance impact on performance. 

The Logits Magnitude demonstrates not only reduced susceptibility to bias $\varepsilon$ during convergence but also a strong correlation with the final classification accuracy, unlike direct computation of magnitudes using vector length. Despite its excellent properties, Logits Magnitude is challenging to directly optimize during training. Furthermore, we aim to explore how to further enhance performance under relatively balanced Logits Magnitude conditions.
% Furthermore, it is necessary to explore how to further evaluate the effectiveness of methods in situations based on the relatively balanced Logits Magnitude condition. 
Therefore, we introduce Regularized Standard Deviation as a novel metric to evaluate methods and overcome the challenges.

\begin{definition}[\textbf{Regularized Standard Deviation}]
Given the logits $\mathbf{z}$, the standard deviation $\sigma(\mathbf{z})$ and logits magnitude $\mathbf{L} $.
The regularized standard deviation $\mathbf{r} \in \mathbb{R}^{K}$ is defined as:
\begin{equation}
    \mathbf{r}_i=\frac{\sigma(\mathbf{z}_i)}{\mathbf{L}_i}.
\end{equation}
\end{definition}

The corresponding $\mathbf{r}$ increases gradually as the classes transition from head to tail shown in Fig.~\ref{fig:Regularized_Standard_Deviation}. This observation is in line with general intuition, as classes with more training samples typically exhibit better feature extraction. Importantly, the computed $\mathbf{r}$ values are quite similar for models derived using different optimization techniques. \textbf{Therefore, the Regularized Standard Deviation can serve as an approximate invariant for further analysis.}

\section{Method}
%For LTR classification across various classes, the number of negative samples is generally uniform, resulting in a more balanced problem. Concurrently, the analysis in Sec~\ref{sec:conjection} finds the stable nature of negative samples. Hence, we can reframe the original task of optimizing classification accuracy on positive samples as a new problem focused on optimizing prediction distribution of negative samples.% thereby adopting a reverse perspective on the problem.
\subsection{Deep Dive into Logits Magnitude} \label{sec:method_analysis}

In classification problems, minimizing the loss $\mathcal{L}$ and maximizing classification accuracy are not completely equivalent objectives.  Given above observation, we derived that the computation of gradients for parameters associated with the "Few" classes is notably influenced by sampling and process perturbations, whereas the impact on "Many" classes is relatively mild. This imbalance will affect classification accuracy. We discuss this process in following.

Suppose achieving relatively balanced magnitudes for each class as we observed in Sec.~\ref{sec:conjection}, denoted as $\mathbf{L}_1 \approx \mathbf{L}_2 \approx ... \approx \mathbf{L}_M$, it is inevitable to encounter variations in the standard deviations $\sigma(\mathbf{z}_i) ={\mathbf{r}_i\mathbf{L}_i}$ across different classes. These variations have a significant impact on the numerical values computed during the training process. \textbf{Although complete elimination of such disparity in $\mathbf{{r_iL_i}}$ raised by the data distribution is not feasible, the impact can be mitigated by significantly reducing the magnitudes $\mathbf{L}$.}

Consider random perturbations $\Delta_i$ that independently affect each class $\mathbf{z}_i$ and are proportional to the standard deviation of their logits, i.e., $\mathbf{z}'_i=\mathbf{z}_i+\Delta_i$ and the perturbation $\Delta_i \sim \xi \mathbf{r}_i \mathbf{L}_i$. Here, $\xi$ is a random variable with an expected value of $\mathbb{E}[\xi]=0$. These perturbations $\Delta$ are commonly introduced during the training process, particularly due to factors such as overfitting in the initial stage of training and bias in the sampling of the training set.

\begin{proposition}
When comparing the use of $\mathbf{z}_i$ and $\mathbf{z}'_i$ for subsequent calculations, the impact is minimal for balanced datasets but can be significant for imbalanced datasets. 
\end{proposition}

The expected value of probability $\mathbf{s}'_i$ can be expressed as:
\begin{equation}
  \mathbb{E}[\mathbf{s}'_i]=\mathbb{E}\left[\frac{1}{\sum_j e^{\mathbf{z}'_j - \mathbf{z}'_i}}\right] = \mathbb{E}\left[\frac{1}{\sum_j e^{(\Delta_j - \Delta_i)} e^{(\mathbf{z}_j-\mathbf{z}_i)}}\right].  
\end{equation}
In the case of balanced datasets, expected with $\mathbf{r}_1 \approx \mathbf{r}_2 \approx \cdots \approx \mathbf{r}_M$, $(\mathbf{r}_j \mathbf{L}_j - \mathbf{r}_i \mathbf{L}_i)$ is close to zero.
\begin{equation}
    % \resizebox{0.9\linewidth}{!}{$
    \mathbb{E}[\mathbf{s}'_i] \approx \frac{1}{\mathbb{E}[\exp{(\Delta_j-\Delta_i)}]}\mathbb{E}\left[\frac{1}{\sum_j e^{\mathbf{z}_j - \mathbf{z}_i}}\right] \\
    = \frac{\mathbb{E}[\mathbf{s}_i]}{\mathbb{E}[\exp{(\Delta_j-\Delta_i)}]}
    % $},
    \label{equ:balanced_bias}
\end{equation}
where $\mathbb{E}[\exp{(\Delta_j-\Delta_i)}]$ can be considered as a constant $\mathcal{E}$. For balanced datasets, the impact of disturbances is consistent across different classes $i$. 
\begin{equation}
    \mathbb{E}[\mathbf{s}'_i]\ / \ \mathbb{E}[\mathbf{s}_i] \approx 1 / \mathcal{E}.
    \label{equ:balanced_bias_approx_e}
\end{equation}
However, for imbalanced datasets, Eq.~\ref{equ:balanced_bias} no longer holds. There is a significant disparity in the values of $\exp{(\Delta_j - \Delta_i)}$ for different class $j$. If $r_j < r_k$, $\mathbb{E}[\exp{(\Delta_j - \Delta_i)}] < \mathbb{E}[\exp{(\Delta_k - \Delta_i)}]$. For instance, when $\xi$ follows a normal distribution we have 
\begin{equation}
    \mathbb{E}[\exp({\Delta_j -\Delta_i})] = \exp({(\sigma(\Delta_j - \Delta_i))^2/2}),
\end{equation}
where $\sigma(\Delta_j - \Delta_i)$ represents the standard deviations of the random variable $(\Delta_j - \Delta_i)$. As a result, the impact of perturbations is directly correlated with the differences in standard deviations between classes, leading to $\mathbb{E}[\mathbf{s}'_i]\ / \ 
 \mathbb{E}[\mathbf{s}_i]$ not consistent across classes $i$, where $\mathbb{E}[\mathbf{s}'_i]\ / \ \mathbb{E}[\mathbf{s}_i]\neq 1/\mathcal{E}$. The expectation $\mathbb{E}[\mathbf{s}_i]$ plays a crucial role as it not only relates to the final predictions but also has a direct correlation with the derivatives obtained from the gradient descent process. Specifically, the expectation of the derivative $\mathbb{E}[\frac{\partial \mathcal{L}}{\partial \mathbf{z_i}}] = \mathbb{E}[(\mathbf{s}_i - \mathbf{y}_i)]$ can be computed as the difference between the expectations $\mathbb{E}[\mathbf{s}_i]$ and $\mathbb{E}[\mathbf{y}_i]$.

Therefore, in the long-tailed scenario, the perturbations $\Delta$ introduced in training process will have a significant impact on model training, ultimately compromising the final classification accuracy.

\textbf{Deliberately reducing the resulting magnitudes holds promise for addressing the effects caused by these perturbations.} Previous methods~\cite{alshammari2022long, kang2019decoupling} can achieve implicit reduction of Logits Magnitude by regularizing the weights of the prediction head. However, these approaches also limits the expressive power of the classification head. Other methods~\cite{cao2019learning,ren2020balanced,cui2019class,menon2020long} also fail to recognize the importance of deliberately optimizing this effective objective. Therefore, we proposed a simple logits retargeting approach (LORT) to directly reduce the Logits Magnitude.

% \red{The perturbations introduced by overfitting in the first stage, bias in training set sampling, and parameter variations during gradient descent all contribute to the aforementioned perturbation, significantly affecting the final classification results. \textbf{Consequently, deliberately reducing the resulting magnitudes can effectively mitigate these influences,} thus improve the model performance. Although certain regularization methods~\cite{alshammari2022long, kang2019decoupling} can reduce logits magnitude $\mathbf{L}_i$, they cannot simultaneously maintain the unconstrained capabilities of $\mathbf{W}$ and preserve relatively consistent magnitudes among classes. Therefore, we proposed a simple logits retargeting approach (LORT) to directly reduce the logits magnitude, satisfying all three desired conditions and providing the model with the necessary prerequisites to achieve exceptional performance.}

\begin{table*}[tb!]
    \centering
    \caption{\textbf{Comparison for CIFAR100-LT, ImageNet-LT and iNaturalist2018 Benchmarks.} Top-1 accuracy (\%)
    is reported and CIFAR100-LT consists of three imbalanced ratio (IR) 100/50/10.
    Our model outperforms all the Baselines and even Upper Bound models on CIFAR100-LT. Furthermore, our method surpasses all baseline methods on both the ImageNet-LT and iNaturalist2018 datasets. Remarkably, our proposed model achieves satisfactory accuracy in the few-class categories. 
    % by means of learning a better classifier and balancing the emphasis on all classes.
    }
    \vspace{-5pt}
    \resizebox{\textwidth}{!}{
    \begin{tabular}{l|c||ccc||cccc||cccc}
      \toprule[1.0pt]
       \multirow{2}{*}{\textbf{\large Method}} & \multirow{2}{*}{\textbf{Reference}} & \multicolumn{3}{c||}{\textbf{CIFAR100-LT}} & \multicolumn{4}{c||}{\textbf{ImageNet-LT}}  & \multicolumn{4}{c}{\textbf{iNaturalist2018}}    \\ 
       \cline{3-13}
        & & IR=100 & IR=50 & IR=10 & Many & Medium & Few & All & Many & Medium & Few & All    \\ 
       \midrule[0.6pt]

       \rowcolor{lightgray}
       \multicolumn{13}{c}{Baseline Methods} \\
       CE & --- & 38.3 & 43.8 & 55.7  & \textbf{65.9} & 37.5 & 7.7 & 44.4 & \textbf{72.2} & 63.0 & 57.2 & 61.7    \\
       Focal~\cite{lin2017focal} & ICCV17 & 38.4 & 44.3 & 55.7 & 36.4 & 29.9 & 16.0 & 30.5 & --- & --- & --- & 61.1  \\  
        LDAM-DRW~\cite{cao2019learning} & NIPS19 & 42.0 & 46.6 & 58.7 & --- & ---& --- & 48.8 & --- & --- & --- & ---\\
       cRT~\cite{kang2019decoupling} & ICLR19 & --- & --- & --- & 61.8 & 46.2 & 27.3 & 49.6 & 69.0 & 66.0 & 63.2 & 65.2\\
       $\tau$-norm~\cite{kang2019decoupling} & ICLR19 & 47.7 & 52.5 & 63.8 & 59.1 & 46.9 & 30.7 & 49.4 & 65.6 & 65.3 & 65.5 & 65.6 \\ 
       
       CB-CE~\cite{cui2019class} & CVPR19 & 39.6 & 45.3 & 58.0
        & 39.6 & 32.7 & 16.8 & 33.2 & 53.4 & 54.8 & 53.2 & 54.0
       \\

       De-confound\cite{tang2020long} & NIPS20 & 44.1 
       & 50.3 & 59.6 & 62.7 & 48.8 & 31.6 & 51.8 & --- & --- & --- & ---\\

       BALMS\cite{ren2020balanced} & NIPS20 & 50.8  & --- & 63.0 & 61.1 & 48.5 & 31.8 & 50.9 & 65.5 & 67.5 & 67.5 & 67.2 \\
       BBN~\cite{zhou2020bbn} & CVPR20 & 42.6  & 47.0 & 59.1 & --- & --- & --- & --- & 49.4 & 70.8 & 65.3 & 66.3\\
       LogitAjust\cite{menon2020long} & ICLR21 & 42.0 & 47.0 & 57.7 & --- & ---& --- & 51.1 & --- & ---& --- & 69.4\\

       DisAlign \cite{zhang2021distribution} & CVPR21 & --- & --- & --- &
       61.3 & 52.2 & 31.4 & 52.9 & 69.0 & \textbf{71.1} & 70.2 & 70.6 \\
       LTWB~\cite{alshammari2022long}& CVPR22 & 53.4 & 57.7 & 68.7 & 62.5  & 50.4 & 41.5 & 53.9 & 71.2 & 70.4 & 69.7 & 70.2\\
       
       AREA~\cite{chen2023area} & ICCV23 & 48.8 & 51.8 & 60.1 & ---  & --- & --- & 49.5 & --- & --- & --- & 68.4 \\

       RBL~\cite{peifeng2023feature} & ICML23 & 53.1 & 57.2 & --- & 64.8  & 49.6 & 34.2 & 53.3 & --- & --- & --- & --- \\

        EWB-FDR~\cite{hasegawa2023exploring} & ICLR24 & 53.0 & --- & --- & 63.4 & 50.0 & 35.1 & 53.2 & --- & --- & --- & --- \\

       \textbf{LORT (ours)} & & \textbf{54.9} & \textbf{58.8} & \textbf{69.7} & 63.2 & \textbf{50.7} & \textbf{42.3} & \textbf{54.4} & 69.2 & 70.7 & \textbf{71.3} & \textbf{70.8}\\
       
       \midrule[0.6pt]
       \rowcolor{lightgray} \multicolumn{13}{c}{Upper Bound Methods} \\

       RIDE~\cite{wang2020long} & ICLR20 & 49.1 & --- & ---  & 67.9 & 52.3 & 36.0 & 56.1 & 66.5 & 72.1 & 71.5 & 71.3\\
       SSD~\cite{li2021self} & ICCV21 & 46.0 & 50.5 & 62.3 & 66.8 & 53.1 & 35.4 & 56.0 & --- & --- & --- & 71.5\\
       PaCo~\cite{cui2021parametric} & ICCV21 & 52.0 & 56.0 & 64.2 & 63.2 & 51.6 & 39.2 & 54.4 & 69.5 & 72.3 & 73.1 & 72.3 \\ 
       OPeN~\cite{zada2022pure} & ICML22 & 51.9 & 56.7 & --- & --- & --- & --- & 55.1 & --- & --- & --- & ---\\  
       BCL~\cite{zhu2022balanced} & CVPR22 & 51.0 & 54.9  & 64.4 & 65.3 & 53.5 & 36.3 & 55.6 & 69.5 & 72.4 & 71.7 & 71.8 \\
      \bottomrule[1.0pt]
    \end{tabular}
    }
    \vspace{-10pt}
    \label{tab:Image_iNa_bench}
\end{table*}

\subsection{Logits Retargeting}
Based on the above analysis, we can observe that when the value of $s_i$ is larger, the impact of perturbation is smaller, and the differences between classes are also smaller. So, an intuitive idea is to make $s_i$ as large as possible during the process.

% Considering the scenario of optimizing negative samples, 
In our method, we divide the original one-hot label into small true label probabilities and large negative label probabilities, distributing them to each class on average. It is essential to ensure that the probability of the positive class is higher than the probability of each negative class for model convergence and discriminability in the true class.

Suppose the number of the categories is $K$. Given the input image $\mathbf{x}$ and label $y$, feature extractor backbone $f(\cdot)$ and classifier $\theta = (\mathbf{W}, \mathbf{b})$, the formulation of our simple logits retargeting method (LORT) is as follows:
\begin{equation}
\begin{aligned}
    & \mathcal{L}(\mathbf{W}, \mathbf{b}; f(\mathbf{x}), y)  = \sum_{i = 1}^{K} -\mathbf{\tilde{y}}_i \cdot \log \left(\frac{e^{\mathbf{z}_i}}{\sum_j e^{\textbf{z}_j}}\right), \\ 
    \mathbf{\tilde{y}}_i  = & \left\{ 
    \begin{array}{ll}
      1 - \delta + \delta / K & \text{if }  i = y\\
      \delta / K, & \text{if } i \neq y
    \end{array}
    \right. \text{ and  }\  \mathbf{z}_i = \mathbf{W}_i^\top f(\mathbf{x}) + \mathbf{b},
\end{aligned}
\end{equation}
where $\delta \in [ 0, 1)$ is a constant value to control the overall negative class probability. Interestingly the above formulation is fundamentally consistent with the conventional Label Smoothing approach~\cite{szegedy2016rethinking} and we can name the hyper-parameter $\delta$ as "label smooth value". \textbf{The primary distinction lies in the negative probability $\delta$,  which can be set as a quite large value in our approach}.

To our best knowledge, Label Smoothing is typically employed in a lightweight manner. For example, $\delta$ is usually set as $0.2$. Our method represents a groundbreaking and unprecedented attempt in LTR field, setting $\delta$ as $0.98$ or $0.99$ to achieve SOTA results.
As theoretically
analysis in Section~\ref{sec:method_analysis}, the larger label smooth value can reduce the Logits Magnitude to obtain better performance. From this perspective, $\delta$ approaching $1$ can yield the best results. However, our method essentially acts as a auxiliary regularization technique to mitigate biases among different class samples and assist the model in convergence. The model itself needs to exhibit both convergence and discriminability, where the ground truth labels should be distinct from the other labels. We believe this represents a trade-off. We will demonstrate experimentally the phenomenon in Sec~\ref{sec:AblationStudy}.

Compared to previous approaches, LORT directly focus on optimizing the fundamental objective Logits Magnitude. In addition to achieving outstanding performance in experiments, the advantages of our method lie in its ease of implementation and its independence from prior information about the number of classes. This represents a comprehensive improvement in both analysis and performance over existing methods.
\section{Experiments}

We conduct experiments on long-tailed image classification benchmarks: CIFAR100-LT \cite{krizhevsky2009learning}, ImageNet-LT \cite{liu2019large} and iNaturalist2018 \cite{van2018inaturalist}. % For the experiments under the same settings, we quoted the results from original papers.  
All experiments are implemented using PyTorch and run on GeForce RTX 3090 (24GB) and A100-PCIE (40GB). Source code will be made publicly available.

\subsection{Datasets and Implementation Details}
We define the class number as $K$ and imbalanced ratio (IR) of the long-tailed datasets as IR = $n_{max} / n_{min}$, where $n_{max}$ and $n_{min}$ is the number of training samples in the largest and smallest class, respectively. 
% Following \cite{cao2019learning}, the long-tailed version CIFAR100-LT was constructed by exponentially decaying sampling with three imbalanced factor IR as 10/50/100. ImageNet-LT is built in \cite{liu2019large} from ImageNet dataset \cite{russakovsky2015imagenet} with IR as 256. iNaturalist2018 \cite{van2018inaturalist} is a real-world, large-scale dataset for species identification with IR as 500.  
In line with the SOTA~\cite{alshammari2022long}, we use ResNet34~\cite{he2016deep} in CIFAR100-LT, ResNeXt50~\cite{xie2017aggregated} in ImageNet-LT and ResNet50~\cite{he2016deep} in iNaturalist2018. For label smooth value, we use $0.98$ in CIFAR100-LT and $0.99$ in ImageNet-LT and iNaturalist2018. 
We report the best result in five trails. Results in Tab.\ref{tab:long-tailed-loss} for CIFAR100LT with IR=100 also follow the same implementation. More details of the datasets and implementation are included in Appendix~\ref{app:details}.

\textbf{Evaluation Metrics.} We evaluate the models on corresponding balanced test dataset and report top-1 accuracy. In line with the protocol in literature \cite{liu2019large}, we give accuracy on three different splits of classes with varying numbers of training data: many (over 100 images), medium (20$\sim$100 images), and few (less than 20 images). 

\begin{figure}[t]
    \centering 
     \vspace{-10pt}
     \includegraphics[width=0.9\linewidth]{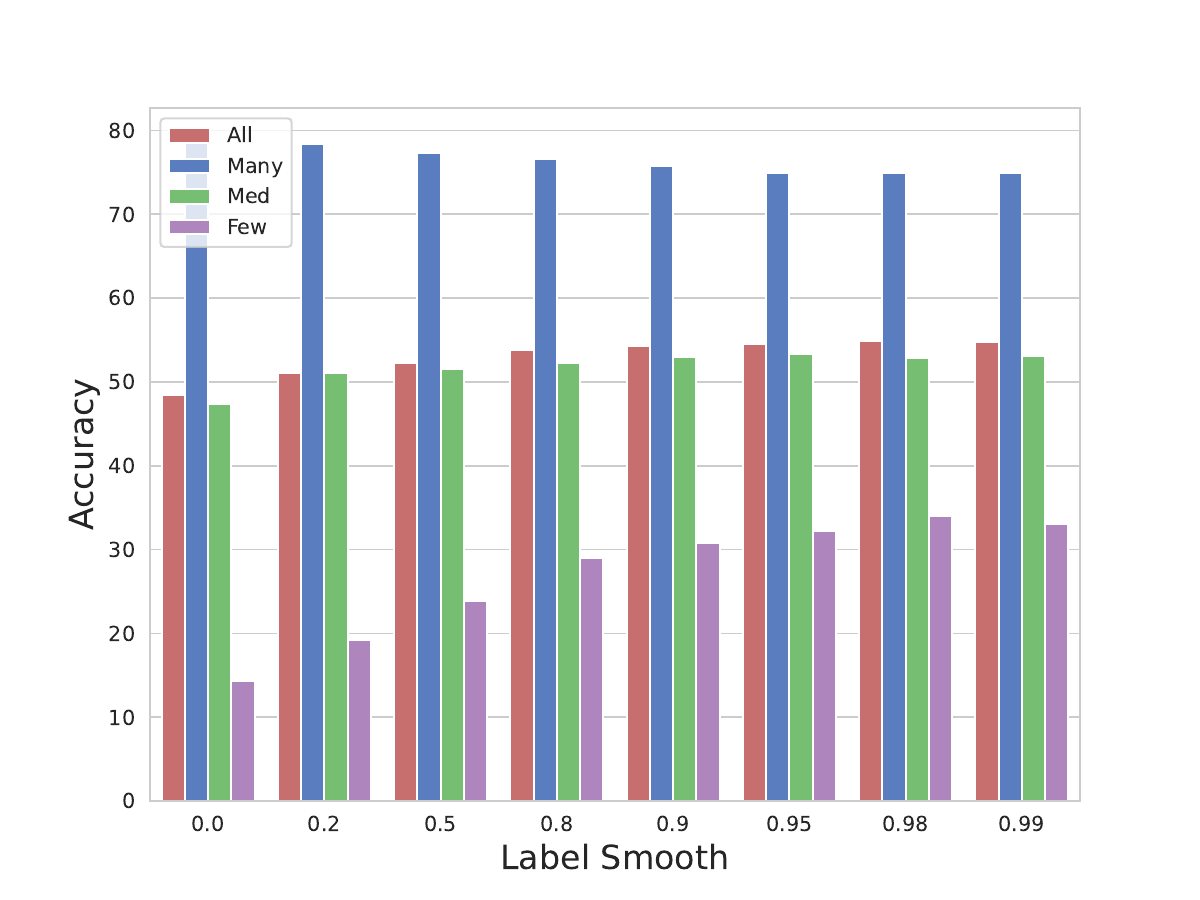}
    \vspace{-10pt}
    \caption{\textbf{Effect of Label Smooth Value.} Tuning smooth value can significantly influence the performance. The overall accuracy will be improved with the growth of smooth value. Emphasis on negative samples probability can effectively enhance the performance of the minority while also improving overall performance.}
    \vspace{-10pt}
    \label{fig:EffectLabelSmooth}
\end{figure}

\subsection{Benchmark Results}
\textbf{Compared Methods.} 
%With the rapid development in the Long-Tailed Recognition field, various methods have been proposed. 
We selected the popular methods in LTR and split them into two categories. (i) Baseline models: Some of these models modify the loss directly, such as cross entropy (CE) loss, class balanced CE loss (CB-CE)~\cite{cui2019class} or focal loss~\cite{lin2017focal}. RBL~\cite{peifeng2023feature} focuses on the perspective of feature learning.
De-confound~\cite{tang2020long}, LogitAjust~\cite{menon2020long},  BALMS~\cite{ren2020balanced}, AREA~\cite{chen2023area} correct the classifier from different analysis aspects. 
% In the one-stage training family, 
BBN~\cite{zhou2020bbn}, LDAM-DRW~\cite{cao2019learning} leverage a learning rate annealing strategy to balance the feature learning process and the re-weighting procedure. The more straightforward method is decoupled learning~\cite{kang2019decoupling}, consisting of cRT~\cite{kang2019decoupling}, $\tau$-norm~\cite{kang2019decoupling}, DisAlign~\cite{zhang2021distribution} and LTWB~\cite{alshammari2022long}.
Among them, LTWB is the most competitive model, whose main contribution is to use simple regularization to learn better robust features in the first stage. EWB-FDR~\cite{hasegawa2023exploring} delved into the success of LTWB by exploring Fisher’s Discriminant Ratio in depth and streamlining the process, albeit at the expense of reduced performance.
 (ii) Upper Bound models: These models tend to make complex design or need more computational resources. 
 % We listed them as the upper bound. 
 They include self-supervised pretraining method PaCo~\cite{cui2021parametric}, BCL~\cite{zhu2022balanced}, additional augmentation training data method OPeN~\cite{zada2022pure}, multi-experts ensemble method RIDE~\cite{wang2020long}, transfer learning method SSD~\cite{li2021self}. We utilize the results reported in these original works.
 % These models tend to learn better feature representations, and we include them as a reference for an upper bound. 

\textbf{Results and Analysis.} Tab.~\ref{tab:Image_iNa_bench} presents the results for long-tailed dataset recognition across three benchmarks. Our method LORT achieves state-of-the-art performance on all the benchmarks. Compared to the origin SOTA model LTWB~\cite{alshammari2022long} on CIFAR100-LT, LORT achieves an improvement of $1\% \sim 1.5\%$ in CIFAR100-LT dataset, $0.5\%$ in ImageNet-LT and $0.6\%$ in iNaturalist2018. In the long-tail tasks without bells and whistles, the improvement is significant.
% Apart from this, Tab.~\ref{tab:long-tailed-loss} also shows the effect of different classifier finetuning strategies when IR=100 under the same configuration. Our method outperforms all the current existing ones. % and shows better stability than the Class-Balanced Loss \cite{cui2019class}  shown in Fig.\ref{fig:SensitivetyLRWD} \textbf{in ablation study}. 
On ImageNet-LT, LORT achieves a significant improvement in classification accuracy for minority classes compared to previous methods, including upper bound models. However, it should be noted that the upper bound models still achieve better overall accuracy on ImageNet-LT and iNaturalist2018 datasets. We attribute this phenomenon to a general feature learning gap in challenging datasets. On CIFAR100-LT, even naive models can perfectly overfit to the imbalanced training set and the core objective is to make the feature general and robust, and finetune the classifier more adaptable to balanced dataset. Hence, LORT can significantly outperform other methods. For large datasets, there is still substantial room for improvement in the representation aspect of the backbone network.
Therefore, advanced methods have a significant advantage in this regard.

\begin{figure}[t]
    \centering 
     \vspace{-10pt}
     \includegraphics[width=0.9\linewidth]{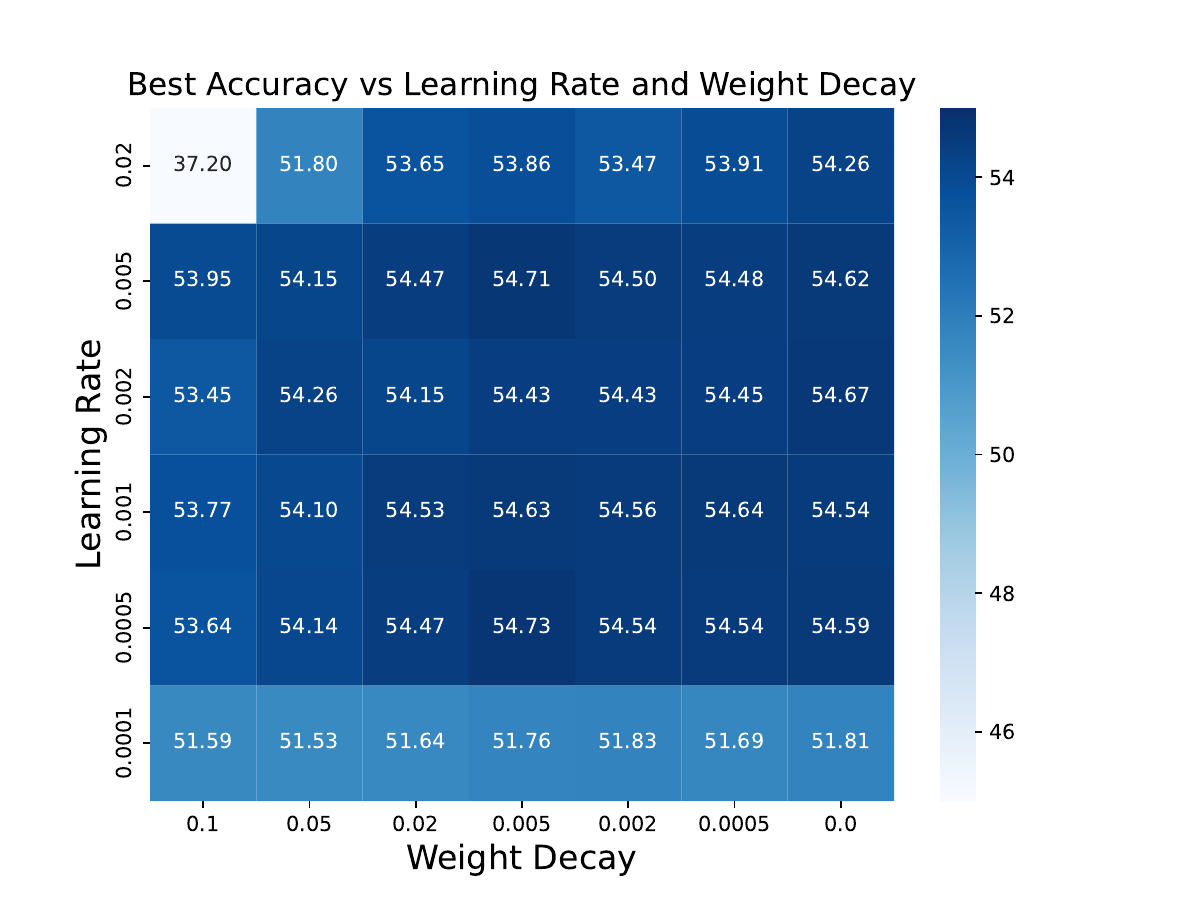}
    \vspace{-10pt}
    \caption{\textbf{Sensitivity Comparison of Learning Rate and Weight Decay}. Tuning learning rate (LR) and weight decay (WD) under the same finetuning epoch budget can have a significant impact on  accuracy. Our method exhibits the strong stability across different LR and WD combinations.}  
    \vspace{-10pt}
    \label{fig:SensitivetyLRWD_ours}
\end{figure}

\subsection{Ablation Study} \label{sec:AblationStudy}

We further conduct ablation study on three aspects: 1) Effect of label smooth value, 2) Sensitivity of learning rate and weight decay, 3) Improvement for different backbone. All experiments are on CIFAR100-LT with IR=$100$.

\textbf{Effect of Label Smooth Value.} The label smooth value is the only adjustable parameter in LORT. In Sec.~\ref{sec:method_analysis}, we suggest using a large label smooth value, and experiments investigating the effect of different values can provide further insights. We evaluate LORT using a range of label smooth values. The value $0.0$ represents the Vanilla CE loss and the value $0.2$ represents the common used label smooth method~\cite{szegedy2016rethinking} which uses soft labels to reduce overfitting. Fig.~\ref{fig:EffectLabelSmooth} shows overall accuracy improvement with the increase of smooth value.
%Given a fixed feature representation, adjusting the classification head is akin to adjusting the decision boundary, which inevitably entails a trade-off between correctly classifying majority class samples and those of other classes. The larger smooth label value can significantly refine the classification boundary of other classes with a lesser sacrifice of correctly classifying majority class samples, where the trade-off is worthwhile. 
The best overall accuracy is achieved with $0.98$ and the adjacent accuracy is close, within the allowable range of disturbance. \textbf{Therefore, we can choose the large label smooth value reassuringly.}

\textbf{Sensitivity of Learning Rate and Weight Decay.} The adjustable hyper-parameters in classifier finetuning are learning rate and weight decay. We use grid search to find the best accuracy and report the sensitivity using heat map. According to Fig.~\ref{fig:SensitivetyLRWD_ours}, it is evident that our method is robust and can achieve a result close to the optimal solution under different combinations. \textbf{Therefore, our proposed model 
exhibits stability and ease of convergence.}

\begin{table}[t!]
    \renewcommand{\arraystretch}{1.3}
    \centering
    \caption{\textbf{Improvement for different feature representation backbones} (in different rows) on CIFAR100-LT with IR=100. Training a backbone by different methods, LORT outperforms all finetuning methods (across columns).}
    \vspace{-10pt}
    \resizebox{\linewidth}{!}{%
    \begin{tabular}{l|ccccccc}
        \toprule
            % \multirow{2}{*}{\textbf{Backbone }} 
            \textbf{Backbone} & \multicolumn{6}{c}{\textbf{Classifier Finetune Methods}}     \\ 
            \cline{2-8} 
            \textbf{Training} & Vanilla &  CE & Focal & LDAM & BS & CB-BCE & LORT(ours)\\
            \hline
            % \rowcolor{lightgray} \multicolumn{7}{c}{CIFAR100-LT IR=100} \\
            CE & 41.59 & 41.67(+0.08) & 44.18(+2.59) & 43.87(+2.28) & 45.03(+3.44) & 45.01(+3.42)   & \textbf{45.18(+3.59)}\\
            \hline
            CE-DRW & 42.38 & 42.38(+0.00) & 42.55(+0.17) & 42.41(+0.03) & 44.27(+1.89) & 44.39(+2.01) & \textbf{44.88(+2.50)}\\
            \hline
            LDAM-DRW & 44.43 & 44.45(+0.02) & 44.58(+0.15) & 44.47(+0.04) & 45.34(+0.91) & 45.52(+1.09) & \textbf{45.86(+1.43)} \\
            \hline
            BS & 45.90 & 45.91(+0.01) & 46.04(+0.14) & 45.90(+0.00) & 46.81(+0.91) & 46.76(+0.86) & \textbf{47.08(+1.28)}\\
            % RIDE(3 experts) \\
            % \hline
            % BCL \\
            % \rowcolor{lightgray} \multicolumn{7}{c}{Imagenet-LT} \\
            % CE \\
            % \hline
            % BCL \\
            % \hline
            % Oracle \\
            \bottomrule
    \end{tabular}%
    } 
    \vspace{-10pt}
    \label{tab:Improvement}
\end{table}
 
\textbf{Improvement for Different Backbone.} Our method has achieved excellent results based on strong feature representation, and we are interested in exploring the potential for improving models with arbitrary representations. Therefore, we first train traditional methods, freeze the trained backbone, and then finetune the classifier using different finetuning methods. In Tab.\ref{tab:Improvement}, our method exhibits the strongest ability to improve performance compared to the other fine-tuning methods evaluated. We find that the first-stage features without prior imbalanced information are more malleable, and different methods can obtain significant improvements, while imbalanced feature representations can not often be effectively improved. Interestingly, the BS finetuning method applied to a BS-trained backbone can still result in significant improvements, indicating the importance of second stage training. \textbf{As a plug-and-play adjustment method, our approach is capable of effectively improving the performance of other models by finetuning the classifier.}
\vspace{-5pt}
\section{Conclusion}
\vspace{-5pt}

Based on a critical analysis of existing classifier retraining methods, we introduce two novel metrics, namely "Logits Magnitude" and "Regularized Standard Deviation". The utilization of these metrics offer fresh insights into model performance and shed light on the key requirements for achieving enhanced accuracy.  Consequently, we propose a simple logits retargeting approach (LORT), achieving both smaller and more balanced  absolute Logits Magnitude by directly setting the optimization targets of logits. Extensive experiments show that LORT obtains achieves state-of-the-art (SOTA) performance on various LTR datasets.

\vspace{2mm}
\noindent\textbf{Limitations}: 
% In this paper, some conclusions are experimental in nature. 
Through extensive experimental exploration and investigation of numerous methods, we have formulated some conclusions in the paper. While these findings are somewhat aligned with intuitive understanding, they are not substantiated by theoretical guarantees, remaining room for further refinement.

% Even though they  align with intuition to a certain extent, they lack theoretical guarantees and there remains room for further refinement.

{
    \small
    \bibliographystyle{ieeenat_fullname}
    \bibliography{main}
}

\clearpage

\appendix

\section{Proof of Proposition 1 Bias Convergence}
\label{app:proof}

\begin{proof}
The loss function is

\begin{equation}
    \mathcal{L}  = \sum_i - \mathbf{y}_i \cdot \log \left(\frac{e^{\mathbf{z}_i}}{\sum_j e^{\mathbf{z}_j}}\right) \\ 
\end{equation}

Using the Chain Rule to calculate derivatives, 

\begin{equation}
    \begin{aligned} 
    \frac{\partial \mathcal{L}}{\partial \mathbf{z}_i} &= \sum_j \frac{\partial \mathcal{L}}{\partial  \mathbf{s}_j}\frac{\partial \mathbf{s}_j}{\partial \mathbf{z}_i} \\
    & = \frac{\partial \mathcal{L}}{\partial \mathbf{s}_i}\frac{\partial \mathbf{s}_i}{\partial \mathbf{z}_i} + \sum_{j\neq i} \frac{\partial \mathcal{L}}{\partial \mathbf{s}_j}\frac{\partial \mathbf{s}_j}{\partial \mathbf{z}_i}\\
    & = -\frac{\mathbf{y}_i}{\mathbf{s}_i}(\mathbf{s}_i(1 - \mathbf{s}_i))-\sum_{j\neq i}\frac{\mathbf{y}_j}{\mathbf{s}_j}(-\mathbf{s}_i \mathbf{s}_j)\\
    & = \mathbf{y}_i(\mathbf{s}_i-1)+\sum_{j\neq i} \mathbf{y}_j \mathbf{s}_i\\
    & = \mathbf{s}_i - \mathbf{y}_i
    \end{aligned}
    \label{equ:derivatives}
\end{equation}

Therefore, the Hessian matrix of $\mathcal{L}$ with respect to $\mathbf{b}$ as $\mathbf{H}$ is as follows:
\begin{equation}
    \mathbf{H}_{ij}=
    \begin{cases}
        \mathbf{s}_i(1 - \mathbf{s}_i) &\text{if $i = j$} \\
        -\mathbf{s}_i \mathbf{s}_j &\text{if $i\neq j$}
    \end{cases}
\end{equation}

And for any vector $\mathbf{x}$, 
\begin{equation}
\mathbf{x}\mathbf{H}\mathbf{x}^\top = \sum_i \mathbf{s}_i \mathbf{x}_i^2 - (\sum_i \mathbf{s}_i \mathbf{x}_i)^2
\end{equation}

Since $g(\mathbf{x}) = \mathbf{x}^2$ is convex function, with $\mathbf{s}_i>0, \sum \mathbf{s}_i = 1$, by Jensen Inequality,
\begin{equation}
    \begin{aligned}
        \sum_i \mathbf{s}_i \mathbf{x}_i^2 &\geq (\sum_i \mathbf{s}_i \mathbf{x}_i)^2, \quad \\ 
        \mathbf{x}\mathbf{H}\mathbf{x}^\top & \geq 0
    \end{aligned}
\end{equation}

Therefore, the Hessian matrix $\mathbf{H}$ is positive semi-definite matrix.
\end{proof}

\newpage

\section{Datasets and Implementation Details} \label{app:details}
We define $K$ as class number and imbalanced ratio (IR) of the long-tailed datasets as IR = $n_{max} / n_{min}$, where $n_{max}$ and $n_{min}$ is the number of training samples in the largest and smallest class, respectively. 
% The detailed information of three datasets and experimental implementation is as follows.

\textbf{Datasets.} 1) CIFAR100-LT: The original balanced  CIFAR100 \cite{krizhevsky2009learning} consists
of 50,000 training images and 10,000 test images of size 32$\times$32 with 100 classes. Following \cite{cao2019learning}, the long-tailed version CIFAR100-LT was constructed by exponentially decaying sampling for the training images of each class, maintaining the balanced test set unchanged simultaneously. Setting the imbalanced factor IR to 10/50/100, we create three long-tailed training datasets as \cite{alshammari2022long}. 2) ImageNet-LT is built in \cite{liu2019large} from ImageNet dataset \cite{russakovsky2015imagenet} with 1000 classes. The number of the images for each class ranges from 5 to 1280, which means the imbalanced factor IR$=$256. 3) iNaturalist2018  \cite{van2018inaturalist} is a real-world, large-scale dataset for species identification of animals and plants with 438K images for 8142 classes. The imbalanced factor is extremely large as 500. 

% \textbf{Evaluation Metrics.} We evaluate the models on the corresponding balanced test/validation dataset and report top-1 accuracy. Following \cite{liu2019large}, we also give accuracy on three different splits of classes with varying numbers of training data: many (over 100 images), medium (20$\sim$100 images), and few (less than 20 images).
        
\textbf{Implementation.} In line with the SOTA method~\cite{alshammari2022long}, we use ResNet34~\cite{he2016deep} in CIFAR100-LT, ResNeXt50~\cite{xie2017aggregated} in ImageNet-LT and ResNet50~\cite{he2016deep} in iNaturalist2018. For each model, we use SGD optimizer with momentum 0.9 and cosine learning rate scheduler~\cite{loshchilov2016sgdr} from given initial learning rate to zero. We use Decoupled Training Scheme which consists of training the feature extractor and finetune the classifier separately. The feature extractor is trained with cross-entropy (CE) loss and suitable weight decay follows \cite{alshammari2022long} and the    classifier is finetuned using our proposed method. On CIFAR100-LT,
We train each model for 200 epochs, with batch size 64 and initial learning rate 0.01. On ImageNet-LT / iNaturalist2018, we train for 200 epochs, with batch size as 128 / 256 and initial learning rate 0.03 / 0.1. 
We use random left-right flipping and cropping as our training augmentation, left-right flipping as testing augmentation for the three datasets with the same configuration  as~\cite{alshammari2022long}. In ImageNet-LT, we also use the color jitter following~\cite{cui2022reslt}. For label smooth value, we use $0.98$ in CIFAR100-LT and $0.99$ in ImageNet-LT and iNaturalist2018. The classifier finetune epoch is 20 with adequate learning ratio and weight. We report the best result in five trails. Results in Tab.~\ref{tab:long-tailed-loss} for CIFAR100LT with IR=100 also follow this implementation.

% \section{Broader Impacts and Limitations}
% \paragraph{Broader Impacts} Our work offers valuable insights for future research. In the realm of long-tailed distributions, we have designed a simple yet effective method from the perspective of logits optimization, which significantly improves the performance of the model. Several promising avenues for future exploration exist, such as incorporating category information priors and designing more complex target logit distribution. In terms of specific method design, LORT explores counter-intuitive perspective to traditional Label Smooth techniques and offers reasonable analyses. This can open up avenues for further in-depth research across various domains apart from long-tailed field.

% \paragraph{Limitations} Throughout our research process, our method fuses theoretical derivations with empirical findings. Some key assumptions and conclusions rest on extensive experimental data, yet they lack rigorous formal proofs. While our method demonstrates strong performance on these three commonly used datasets, we cannot assure that these empirical conclusions will persistently hold true when implemented in complex real-world scenarios. We earnestly hope that researchers can provide a more profound theoretical elucidation for these observed phenomena in future works.

\end{document}